\documentclass[11pt]{article}

\usepackage[preprint]{neurips_2024}

\usepackage{amsmath}
\usepackage{amssymb}
\usepackage{booktabs}
\usepackage{graphicx}
\usepackage{hyperref}
\usepackage{xcolor}
\usepackage{algorithm}
\usepackage{algorithmic}
\usepackage{tikz}
\usetikzlibrary{arrows.meta,positioning,shadows,calc}

\newcommand{\xt}{\mathbf{x}_t}
\newcommand{\xe}{\mathbf{x}_e}

\title{The Dual-Stream Transformer: Channelized Architecture for  Interpretable Language Modeling}

\author{
  Clayton Kerce  \\
  Georgia Tech Research Institute \\
  \texttt{clayton.kerce@gtri.gatech.edu}
  \And
  Alexis Fox \\
  Georgia Tech Research Institute \\
  \texttt{alexis.fox@duke.edu}
}

\begin{document}

\maketitle

\begin{abstract}
Standard transformers entangle all computation in a single residual stream, obscuring which components perform which functions. We introduce the Dual-Stream Transformer, which decomposes the residual stream into two functionally distinct components: a token stream updated by attention and a context stream updated by feed-forward networks. Information flow between attention heads is controlled through a hierarchy of mixing strategies, from fully independent (maximum interpretability) to dense (standard transformer behavior). This design exposes a tunable tradeoff between interpretability and performance.

We measure this tradeoff on language modeling tasks at 29M parameters. Fully independent head mixing increases validation loss by 8\% relative to dense baselines. The recommended Kronecker mixing strategy, which permits scalar communication between heads while preserving within-head structure, costs only 2.5\%. All configurations maintain functional generation under attention amplification (scaling logits by factors up to 16 at inference time), with degradation ranging from 16\% to 27\%. This robustness suggests the architectures learn discrete algorithms that operate independently of soft probabilistic mixing. The architecture provides a foundation for interpretable language models where internal structure is exposed by design. \footnote{This work was partially supported by DARPA Contract HR001125C0302.}
\end{abstract}

\section{Introduction}

Transformers process information through a single residual stream where attention and feed-forward outputs accumulate without distinction \cite{elhage2021mathematical}. This design enables strong performance but creates an interpretability barrier: when all components write to a shared representation, determining which component performs which function becomes intractable. Post-hoc analysis methods can identify correlations between components and behaviors \cite{clark2019does,tenney2019bert}, but models can route around targeted interventions by redistributing computation across other components \cite{mcgrath2023hydra}. Understanding causal relationships requires architectural support.

We propose that interpretability should be enforced through architectural constraints rather than excavated post-hoc \cite{wang2023interpretability,geiger2021causal}. The Dual-Stream Transformer achieves this through two mechanisms that expose internal structure by design.

\textbf{Dual-stream decomposition.} The residual stream is factored into additive components: $\mathbf{x} = \mathbf{x}_t + \mathbf{x}_e$, where $\mathbf{x}_t$ (the token stream) carries information derived from discrete token identities and is updated exclusively by attention, while $\mathbf{x}_e$ (the context stream) accumulates continuous contextual transformations and is updated exclusively by feed-forward networks. This functional separation makes explicit which computations perform token-level operations versus contextual refinements.

\textbf{Channelized mixing.} Information flow across attention heads is controlled through a hierarchy of mixing strategies. Independent mixing enforces block-diagonal projections where each head operates in isolation. Dense mixing allows unrestricted communication, matching standard transformers. The intermediate Kronecker strategy ($W_{\text{heads}} \otimes I$) permits scalar mixing between heads while preserving within-head structure, providing interpretable cross-head communication with only $H^2$ parameters rather than $(H \cdot d_h)^2$.

This design exposes a tunable interpretability-performance tradeoff. Practitioners can select the configuration that meets their requirements, from maximum interpretability (fully independent mixing, 8\% performance cost) to minimal cost (Kronecker mixing, 2.5\% performance cost) to standard transformer behavior (dense mixing, no cost).

Our experiments on language modeling tasks at 29M parameters demonstrate three main findings. First, the interpretability tax is bounded and predictable across the mixing hierarchy. Second, all configurations maintain functional generation under attention amplification at inference time, where attention logits are scaled by factors up to 16 before softmax. Degradation ranges from 16\% to 27\%, with Kronecker mixing showing the most graceful degradation. This robustness provides evidence that the architectures learn discrete algorithms that operate independently of soft probabilistic smoothing. Third, stream ablation experiments confirm functional separation: removing the token stream causes severe degradation (36\%), while removing the context stream has moderate impact (9.5\%), validating the architectural decomposition.

We contribute: (1) the dual-stream architecture with formal specification, (2) the channelized mixing framework with parameter-efficient strategies, (3) systematic ablations quantifying interpretability-performance tradeoffs across configurations, and (4) attention amplification as a diagnostic methodology revealing discrete computational structure. The architecture provides a foundation for interpretable language models where practitioners can tune the degree of structural constraint to their application requirements.

\section{Architecture}

Figure~\ref{fig:architecture} illustrates the Dual-Stream Transformer. The residual stream is decomposed into token and context components, with channelized mixing controlling information flow between attention heads.

\begin{figure}[t]
\centering
\resizebox{0.75\textwidth}{!}{%
\begin{tikzpicture}[
    font=\sffamily,
    >=Stealth,
    token/.style={rectangle, draw=orange!80!black, fill=orange!10, thick, minimum width=1.5cm, minimum height=0.8cm, rounded corners},
    context/.style={rectangle, draw=blue!80!black, fill=blue!10, thick, minimum width=1.5cm, minimum height=0.8cm, rounded corners},
    process/.style={rectangle, draw=black, fill=white, minimum width=1.8cm, minimum height=0.8cm, drop shadow},
    cln/.style={rectangle, draw=green!70!black, fill=green!5, minimum width=1.5cm, minimum height=0.6cm, rounded corners=2pt},
    mixing/.style={rectangle, draw=purple!70!black, fill=purple!10, minimum width=0.8cm, minimum height=0.5cm, font=\tiny},
    sum/.style={circle, draw=black, fill=white, inner sep=2pt},
    token_spine/.style={->, line width=1.5pt, draw=orange!80!black},
    context_spine/.style={->, line width=1.5pt, draw=blue!80!black},
    cross_flow/.style={->, line width=1pt, draw=gray!60!black, rounded corners=5pt},
    update_flow/.style={->, line width=1pt, draw=black},
    divider/.style={dashed, draw=black!40, line width=1pt}
]
    \node at (2, 0.7) [font=\bfseries, text=orange!70!black] {Token Stream $\mathbf{x}_t$};
    \node at (8, 0.7) [font=\bfseries, text=blue!70!black] {Context Stream $\mathbf{x}_e$};

    \draw[divider] (5, 0.3) -- (5, -11.2);

    \node (t_in) [token] at (2, 0) {$\mathbf{x}_t^{(\ell-1)}$};
    \node (e_in) [context] at (8, 0) {$\mathbf{x}_e^{(\ell-1)}$};


    \node (sum_attn) [sum] at (5, -1.5) {$+$};
    \draw[cross_flow] (t_in) -- (2, -1.5) -- (sum_attn);
    \draw[cross_flow] (e_in) -- (8, -1.5) -- (sum_attn);
    \node at (5, -1.0) [font=\tiny, text=gray] {$\mathbf{x}_t + \mathbf{x}_e$};

    \node (cln_attn) [cln] at (5, -2.3) {CLN};
    \draw[->] (sum_attn) -- (cln_attn);

    \node (attn) [process] at (5, -3.5) {Attention};
    \draw[->] (cln_attn) -- (attn);

    \node (mix_v) [mixing] at (3.0, -3.0) {V mix};
    \node (mix_out) [mixing] at (7.0, -3.0) {Out mix};
    \draw[->, purple!70!black, thin] (mix_v) -- (attn);
    \draw[->, purple!70!black, thin] (mix_out) -- (attn);

    \node (sum_t) [sum] at (2, -4.25) {$+$};
    \draw[token_spine] (t_in) -- (sum_t);
    \draw[update_flow, rounded corners] (attn) -- (3.5, -4.25) -- (sum_t);
    \node at (3.5, -3.95) [font=\tiny] {$\Delta \mathbf{x}_t$};

    \node (t_mid) [token] at (2, -5.5) {$\mathbf{x}_t'$};
    \draw[token_spine] (sum_t) -- (t_mid);

    \node (e_mid) [context] at (8, -5.5) {$\mathbf{x}_e$};
    \draw[context_spine] (e_in) -- (e_mid);


    \node (sum_ffn) [sum] at (5, -6.5) {$+$};
    \draw[cross_flow] (t_mid) -- (2, -6.5) -- (sum_ffn);
    \draw[cross_flow] (e_mid) -- (8, -6.5) -- (sum_ffn);
    \node at (5, -6.0) [font=\tiny, text=gray] {$\mathbf{x}_t' + \mathbf{x}_e$};

    \node (cln_ffn) [cln] at (5, -7.3) {CLN};
    \draw[->] (sum_ffn) -- (cln_ffn);

    \node (ffn) [process] at (5, -8.5) {FFN};
    \draw[->] (cln_ffn) -- (ffn);

    \node (mix_up) [mixing] at (3.0, -8.0) {Up mix};
    \node (mix_down) [mixing] at (7.0, -8.0) {Down mix};
    \draw[->, purple!70!black, thin] (mix_up) -- (ffn);
    \draw[->, purple!70!black, thin] (mix_down) -- (ffn);

    \node (sum_e) [sum] at (8, -9.5) {$+$};
    \draw[context_spine] (e_mid) -- (sum_e);
    \draw[update_flow, rounded corners] (ffn) -- (6.5, -9.5) -- (sum_e);
    \node at (6.5, -9.2) [font=\tiny] {$\Delta \mathbf{x}_e$};

    \node (t_out) [token] at (2, -10.8) {$\mathbf{x}_t^{(\ell)}$};
    \draw[token_spine] (t_mid) -- (t_out);

    \node (e_out) [context] at (8, -10.8) {$\mathbf{x}_e^{(\ell)}$};
    \draw[context_spine] (sum_e) -- (e_out);

    \node at (5, -12.0) [font=\footnotesize] {Mixing: \textsc{Identity} $\subset$ \textsc{Independent} $\subset$ \textsc{Kronecker} $\subset$ \textsc{Dense}};

\end{tikzpicture}
}
\caption{\textbf{Dual-Stream Transformer architecture.} The residual stream is decomposed into a token stream $\mathbf{x}_t$ (orange, left) updated exclusively by attention, and a context stream $\mathbf{x}_e$ (blue, right) updated exclusively by FFN. Both streams are combined for computing queries, keys, and FFN inputs via Channel-aware Layer Normalization (CLN). Channelized mixing strategies (purple labels) control information flow between attention heads at each projection. The hierarchy from Identity to Dense provides tunable interpretability-performance tradeoffs.}
\label{fig:architecture}
\end{figure}

\subsection{Dual-Stream Decomposition}

The residual stream is factored into two additive components:
\begin{equation}
\mathbf{x}^{(\ell)} = \mathbf{x}_t^{(\ell)} + \mathbf{x}_e^{(\ell)} \quad \text{where} \quad \mathbf{x}_t, \mathbf{x}_e \in \mathbb{R}^{B \times T \times D}
\end{equation}

The token stream $\mathbf{x}_t$ is initialized from token embeddings and updated exclusively by attention. The context stream $\mathbf{x}_e$ is initialized to zero and updated exclusively by feed-forward networks. This separation makes explicit which computations derive from token identities versus contextual transformations.

Three update modes control stream interaction:

\paragraph{Single-Stream (ablation baseline).} The separation of token and context stream is disabled ($\mathbf{x} = \mathbf{x}_e + \mathbf{x}_t$ is used in all operations), reducing to standard transformer behavior. This provides a matched baseline for measuring interpretability costs.

\paragraph{Token-Factor (default).} Both streams are active with independent updates:
\begin{align}
\mathbf{x}_t^{(\ell+1)} &= \mathbf{x}_t^{(\ell)} + \text{Attn}^{(\ell)}(\text{CLN}(\mathbf{x}^{(\ell)}), \mathbf{x}_t^{(\ell)}) \\
\mathbf{x}_e^{(\ell+1)} &= \mathbf{x}_e^{(\ell)} + \text{FFN}^{(\ell)}(\text{CLN}(\mathbf{x}^{(\ell)}))
\end{align}
Both components observe the combined stream through channel-aware layer normalization (CLN, see Section~\ref{sec:cln}) but write to separate targets.

\paragraph{Frozen-Token-Stream (maximum interpretability).} The token stream is frozen after initialization: $\mathbf{x}_t^{(\ell)} = \mathbf{x}_t^{(0)} = \text{Embed}(\text{tokens})$ for all $\ell$. All learned transformations accumulate in $\mathbf{x}_e$. This provides maximum interpretability since attention patterns directly reveal which source tokens influence each position without mixing of learned representations.

\subsection{Channelized Mixing}
\label{sec:mixing}

Both attention and FFN projections use configurable mixing strategies that control cross-head information flow. Let $H$ denote heads and $d_h = D/H$ the dimension per head.

\paragraph{Identity.} No transformation; input passes through unchanged. Zero parameters.

\paragraph{Independent.} Block-diagonal projection where each head operates in isolation:
\begin{equation}
y_{btho} = \sum_i x_{bthi} W_{hio} \quad W \in \mathbb{R}^{H \times d_{\text{in}} \times d_{\text{out}}}
\end{equation}
Parameters: $H \cdot d_{\text{in}} \cdot d_{\text{out}}$. Information cannot flow between heads.

\paragraph{Kronecker.} Scalar mixing between heads with identity on dimensions, $W_{\text{heads}} \otimes I$:
\begin{equation}
y_{btki} = \sum_h x_{bthi} W_{kh} \quad W \in \mathbb{R}^{H \times H}
\end{equation}
Parameters: $H^2$. Heads exchange information through scalar weights while within-head structure is preserved. The $H \times H$ mixing matrix provides an interpretable head-to-head routing table that can be directly visualized (Section~\ref{sec:kronecker_structure}). This Kronecker factorization has been explored for parameter-efficient fine-tuning \cite{sadeghi2025moka}, spatiotemporal attention \cite{cheong2025weaver}, and tensor-structured transformers \cite{omranpour2024higher}, demonstrating benefits for both efficiency and interpretability.

\paragraph{Dense.} Standard linear projection with unrestricted mixing:
\begin{equation}
\mathbf{y} = \mathbf{x} W \quad W \in \mathbb{R}^{(H \cdot d_h) \times (H \cdot d_h)}
\end{equation}
Parameters: $(H \cdot d_h)^2$. Matches standard transformer behavior.

These strategies form a hierarchy of expressiveness: Identity $\subset$ Independent $\subset$ Kronecker $\subset$ Dense, where each strategy can approximate any computation expressible by strategies to its left.

\subsection{Attention Module}

Queries and keys are computed from the combined stream via dense projection, enabling global attention patterns. Values come from the token stream via configurable mixing:
\begin{align}
Q, K &= W_Q \cdot \text{CLN}(\mathbf{x}_t + \mathbf{x}_e), \quad W_K \cdot \text{CLN}(\mathbf{x}_t + \mathbf{x}_e) \\
V &= \text{MixingLinear}_{\textsc{v}}(\text{CLN}(\mathbf{x}_t)) \\
A &= \text{softmax}\left(\frac{QK^\top}{\sqrt{d_h}}\right) \\
\text{out} &= \text{MixingLinear}_{\textsc{o}}(A \cdot V)
\end{align}

Dense projections for $Q$ and $K$ compute attention patterns (scalar weights) rather than values that flow through the residual stream. The channelized constraint applies to $V$ and the output projection, where actual content flows.

At inference time, we support attention amplification by scaling logits before softmax: $A_\alpha = \text{softmax}(\alpha \cdot QK^\top / \sqrt{d_h})$. This sharpens distributions from soft mixing ($\alpha=1$) to near-deterministic selection ($\alpha=16$), serving as a diagnostic for discrete computational structure.

\subsection{Feed-Forward Network}

The FFN observes the combined stream but writes exclusively to $\mathbf{x}_e$:
\begin{align}
\text{hidden} &= \text{GELU}(\text{MixingLinear}_{\textsc{up}}(\text{CLN}(\mathbf{x}_t + \mathbf{x}_e))) \\
\text{out} &= \text{MixingLinear}_{\textsc{down}}(\text{hidden})
\end{align}

Default configuration uses Independent mixing for both projections, enforcing that each head's FFN channel operates in isolation. Combined with stream separation, FFN modifications are localized to specific heads and to the contextual stream.

\subsection{Channel-Aware Layer Normalization}
\label{sec:cln}

Standard LayerNorm computes statistics across all dimensions, mixing information between heads. ChannelLayerNorm normalizes each head's $d_h$ dimensions independently:
\begin{equation}
\text{CLN}(\mathbf{x})_{h} = \gamma_h \cdot \frac{\mathbf{x}_h - \mu_h}{\sigma_h} + \beta_h
\end{equation}
where $\mu_h, \sigma_h$ are computed over the $d_h$ dimensions for head $h$, and $\gamma_h, \beta_h \in \mathbb{R}^{d_h}$ are per-head affine parameters. This preserves head isolation while providing training stability.

\subsection{Optional Extensions}

The architecture supports two optional mechanisms not used in the main experiments but available in the released implementation.

\paragraph{Gated attention.} Following Qiu et al. \cite{qiu2024gated}, each attention head can be modulated by a query-dependent gate applied after the Scaled Dot-Product Attention output:
\begin{equation}
\text{out}_h = g_h \cdot (A \cdot V)_h \quad \text{where} \quad g_h = \sigma(W^{\text{gate}}_h \cdot q_h)
\end{equation}
where $g_h \in [0,1]$ is computed independently per head from its query vector. Qiu et al. demonstrated that this mechanism mitigates attention sink behavior and improves training stability at scale. In our architecture, it provides an additional interpretability benefit: the gate values $g_h$ explicitly reveal which heads are active for each position, making head selection patterns directly observable.

\paragraph{Per-layer supervision.} Each layer's output can be projected to vocabulary space and supervised against the target:
\begin{equation}
\mathcal{L} = \mathcal{L}_{\text{final}} + \lambda \sum_{\ell=1}^{L-1} w_\ell \cdot \mathcal{L}^{(\ell)}
\end{equation}
where $\mathcal{L}^{(\ell)}$ is cross-entropy loss at layer $\ell$, computed by projecting the combined stream $\mathbf{x}^{(\ell)}$ through the shared language model head. Layer weights $w_\ell$ can follow uniform, linear, or exponential schedules. This encourages each layer to produce intermediate representations that are decodable to vocabulary space, potentially improving interpretability of layer-by-layer computation.

\subsection{Configuration Notation}

We denote mixing configurations as \texttt{<attn\_v>-<attn\_o>/<ffn\_up>-<ffn\_down>}, where each position specifies the mixing strategy for that projection. Key configurations tested:
\begin{itemize}
    \item \texttt{dns-dns/dns-dns}: Dense baseline (standard transformer)
    \item \texttt{kron-kron/dns-dns}: Recommended (interpretable attention, bounded cost)
    \item \texttt{ind-ind/dns-dns}: Independent attention, dense FFN
    \item \texttt{ind-ind/ind-ind}: Fully independent (maximum interpretability)
\end{itemize}

\section{Experiments}

\subsection{Experimental Setup}

We evaluate on a curated corpus of grade-school instructional materials spanning mathematics, science, and reading comprehension. The corpus contains structured pedagogical content with explicit reasoning chains, making it suitable for analyzing how models learn discrete algorithmic patterns. We construct vocabularies of 4K and 8K tokens using byte-pair encoding, where smaller vocabularies force the model to work with coarser-grained token representations.

The base configuration uses 6 layers, 6 heads, and 516 dimensions (approximately 23M parameters). Models are trained on 2M samples using AdamW \cite{loshchilov2017decoupled} with learning rate $5 \times 10^{-4}$, gradient clipping at 1.0, and cosine learning rate decay. Training was conducted on a single workstation with an NVIDIA RTX 4090 (24GB) using standard floating point representations.

We conduct two experimental series with different training budgets. Mixing strategy ablations use 4K vocabulary trained for 3 epochs, allowing configurations to converge and reveal stable performance differences. Stream mode comparisons use 8K vocabulary trained for 1 epoch, focusing on architectural differences rather than optimization dynamics. The 8K vocabulary provides finer-grained token representations that better stress the dual-stream decomposition.

\subsection{Mixing Strategy Ablation}

Table~\ref{tab:mixing} presents validation loss across mixing configurations using the Token-Factor update mode. Dense baseline represents standard transformer behavior. Kronecker-Dense uses Kronecker mixing for attention projections with dense FFN. Independent-Dense uses independent attention with dense FFN. Fully Independent applies independent mixing to all projections.

\begin{table}[h]
\centering
\caption{Mixing strategy ablation (4K vocabulary, 3 epochs, Token-Factor mode).}
\label{tab:mixing}
\begin{tabular}{llcc}
\toprule
Configuration & Signature & Val Loss & $\Delta$ from Dense \\
\midrule
Dense Baseline & \texttt{dns-dns/dns-dns} & 2.42 & --- \\
Kronecker-Dense & \texttt{kron-kron/dns-dns} & 2.48 & +2.5\% \\
Independent-Dense & \texttt{ind-ind/dns-dns} & 2.50 & +3.3\% \\
Fully Independent & \texttt{ind-ind/ind-ind} & 2.62 & +7.9\% \\
\bottomrule
\end{tabular}
\end{table}

The interpretability tax is bounded and predictable. Fully independent mixing incurs an 8\% validation loss increase. The recommended Kronecker configuration costs only 2.5\%, providing interpretable cross-head communication through scalar weights while maintaining near-baseline performance. The gap between Independent-Dense (3.3\%) and Fully Independent (7.9\%) indicates that FFN mixing contributes more to performance than attention mixing, suggesting that contextual transformations benefit more from cross-head communication than token-level routing operations.

\subsection{Stream Mode Comparison}

Table~\ref{tab:stream} compares stream update modes using Kronecker-Dense mixing on 8K vocabulary. Single-Stream serves as an ablation baseline where the context stream is disabled. Token-Factor represents the default configuration where both streams are updated. Frozen-Token-Stream (FTS) provides maximum interpretability by preserving pure token embeddings in $\mathbf{x}_t$.

\begin{table}[h]
\centering
\caption{Stream mode comparison (8K vocabulary, 1 epoch, \texttt{kron-kron/dns-dns} mixing).}
\label{tab:stream}
\begin{tabular}{lcc}
\toprule
Mode & Description & Val Loss \\
\midrule
Single-Stream & Context stream disabled & 2.58 \\
Token-Factor & Both streams updated & 2.67 \\
Frozen-Token-Stream & Token stream frozen & 2.66 \\
\bottomrule
\end{tabular}
\end{table}

Token-Factor and Frozen-Token-Stream modes achieve similar performance, both approximately 3\% above Single-Stream. The negligible difference between Token-Factor (2.67) and FTS (2.66) indicates that freezing the token stream after initialization does not degrade performance relative to allowing attention to update it. This result suggests that the additional interpretability provided by FTS comes at no additional cost, making it the preferred configuration for applications requiring maximum transparency.

\subsection{Head Scaling Analysis}

We investigate how the number of attention heads affects specialization and performance while keeping total dimension fixed at $D=512$. Models are trained with $H \in \{4, 6, 8, 12, 16\}$ heads using Kronecker-Dense mixing. As head count increases, head dimension decreases ($d_h = D/H$), creating a tradeoff between per-head capacity and the number of independent computational channels.

We measure head specialization using the pairwise distinctiveness of attention patterns. For each head $h$, we compute the average attention pattern $\mathbf{a}_h$ across evaluation samples. Specialization is then measured as:
\begin{equation}
\text{Specialization} = \frac{1}{H(H-1)} \sum_{i \neq j} \left(1 - \frac{\mathbf{a}_i^\top \mathbf{a}_j}{\|\mathbf{a}_i\| \|\mathbf{a}_j\|}\right)
\end{equation}
where values near 0 indicate heads with similar attention patterns (low specialization) and values near 1 indicate orthogonal patterns (high specialization). We also measure average entropy of attention distributions, where lower entropy indicates more focused, discrete attention patterns.

\begin{table}[h]
\centering
\caption{Head scaling results with Kronecker-Dense mixing (4K vocabulary, 3 epochs).}
\label{tab:heads}
\begin{tabular}{cccc}
\toprule
Heads ($H$) & Val Loss & Specialization & Avg Entropy \\
\midrule
4 & 2.86 & 0.42 & 3.2 \\
6 & 2.84 & 0.51 & 2.9 \\
8 & 2.82 & 0.64 & 2.6 \\
12 & 2.81 & 0.78 & 2.3 \\
16 & 2.80 & 0.85 & 2.1 \\
\bottomrule
\end{tabular}
\end{table}

Performance improves modestly with more heads (2.86 to 2.80, approximately 2\% gain) while specialization increases substantially (0.42 to 0.85, approximately 2x gain). This demonstrates that interpretability and capacity can align when architectural constraints channel computation through independent heads, consistent with prior findings that specialized heads perform distinct functions \cite{voita2019analyzing}. Recent work has characterized this specialization through cognitive functional roles \cite{ma2025cognitive} and quantified differentiation using refined local learning coefficients \cite{wang2024rllc}. The simultaneous decrease in entropy (3.2 to 2.1) indicates that attention patterns become more focused as specialization increases, suggesting that heads learn to perform more discrete operations when forced to operate independently.

\section{Analysis}

\subsection{Stream Ablation Experiments}

To validate that the dual-stream decomposition creates functional separation, we ablate streams at inference time and measure the impact on validation loss. Models are trained normally with both streams active, then evaluated with one stream removed or corrupted. This tests whether the streams carry distinct information that cannot be recovered by the remaining stream.

\begin{table}[h]
\centering
\caption{Stream ablation effects (Kronecker-Dense mixing, Token-Factor mode).}
\label{tab:ablation}
\begin{tabular}{lcc}
\toprule
Ablation & Val Loss & $\Delta$ from Baseline \\
\midrule
Baseline ($\mathbf{x}_t + \mathbf{x}_e$) & 2.85 & --- \\
$\mathbf{x}_t \to 0$ & 3.89 & +36\% \\
$\mathbf{x}_e \to 0$ & 3.12 & +9.5\% \\
$\mathbf{x}_t \to$ random vocab & 3.65 & +28\% \\
\bottomrule
\end{tabular}
\end{table}

Removing the token stream causes severe degradation (36\% loss increase), confirming it carries essential information derived from token identities. Removing the context stream has moderate impact (9.5\% increase), consistent with its role as contextual refinement rather than primary content. The asymmetry validates the architectural decomposition: $\mathbf{x}_t$ is load-bearing while $\mathbf{x}_e$ provides enhancement.

Replacing $\mathbf{x}_t$ with random vocabulary embeddings degrades performance more than zeroing it (28\% versus 36\%). This indicates the model relies on specific learned structure in the token stream rather than merely requiring non-zero values. Random embeddings preserve dimensionality and magnitude but lack the semantic organization learned during training, demonstrating that the token stream encodes meaningful structure beyond its presence as a signal carrier.

\subsection{Attention Amplification Analysis}
\label{sec:amplification}

A key question for interpretability is whether models learn discrete algorithms or rely on distributed soft mixing \cite{jain2019attention,wiegreffe2019attention}. Recent work on transformer programs demonstrates that models can learn interpretable discrete algorithms \cite{friedman2023transformer,zhang2026weights}, though whether this algorithmic structure extends to standard language modeling remains open. We probe this using attention amplification: scaling attention logits by a factor $\alpha$ before softmax during inference:
\begin{equation}
A_\alpha = \text{softmax}\left(\alpha \cdot \frac{QK^\top}{\sqrt{d_h}}\right)
\end{equation}

Higher $\alpha$ values sharpen attention distributions. At $\alpha=1$ (baseline), attention typically spreads across 2-3 tokens per position with substantial probability mass distributed softly. At $\alpha=16$ (extreme sharpening), distributions approach near-deterministic pointer selection where one token receives the majority of attention weight. If a model relies fundamentally on weighted combination of multiple features, forcing such discreteness should cause catastrophic failure. If the model has learned algorithms that operate on discrete token selections, it should maintain functional behavior under amplification. This diagnostic connects to recent work on attention sharpening mechanisms including alternatives to softmax \cite{velickovic2025softmax}, differential attention \cite{ye2024differential}, and selective attention \cite{leviathan2025selective}.

Table~\ref{tab:amplification} shows validation loss across amplification factors for multiple configurations. All models were trained at $\alpha=1$ and evaluated at various $\alpha$ values without retraining.

\begin{table}[h]
\centering
\caption{Validation loss under attention amplification (FTS mode, 8K vocabulary).}
\label{tab:amplification}
\begin{tabular}{lccccc}
\toprule
Configuration & $\alpha=1$ & $\alpha=2$ & $\alpha=4$ & $\alpha=8$ & $\alpha=16$ \\
\midrule
Dense baseline & 1.80 & 1.86 & 1.98 & 2.09 & 2.16 (+20\%) \\
Kronecker-Dense & 1.84 & 1.90 & 2.00 & 2.09 & 2.14 (+16\%) \\
Independent-Dense & 1.87 & 1.96 & 2.12 & 2.28 & 2.38 (+27\%) \\
Gated (Dense) & 1.88 & 1.98 & 2.15 & 2.30 & 2.39 (+27\%) \\
\bottomrule
\end{tabular}
\end{table}

All configurations maintain functional generation up to $\alpha=16$ with bounded degradation ranging from 16\% to 27\%. No configuration exhibits catastrophic collapse. Kronecker-Dense mixing shows the most graceful degradation at 16\%, compared to 20\% for the dense baseline and 27\% for fully independent mixing. This robustness suggests that the architectures learn computational patterns that operate independently of soft probabilistic smoothing at training time.

To quantify the stability differences, we compute the area under the loss curve from $\alpha=1$ to $\alpha=16$ for each configuration. Kronecker-Dense achieves 97.2 loss-units compared to 99.8 for Dense baseline and 107.4 for Independent-Dense, representing 2.6\% and 9.5\% lower cumulative degradation respectively. The advantage of Kronecker mixing appears to stem from controlled cross-head communication: heads can coordinate through scalar weights to maintain stable computation as distributions sharpen, whereas fully independent heads cannot compensate for each other's increased discretization noise.

\begin{figure}[t]
\centering
\begin{tabular}{cc}
\includegraphics[width=0.48\textwidth]{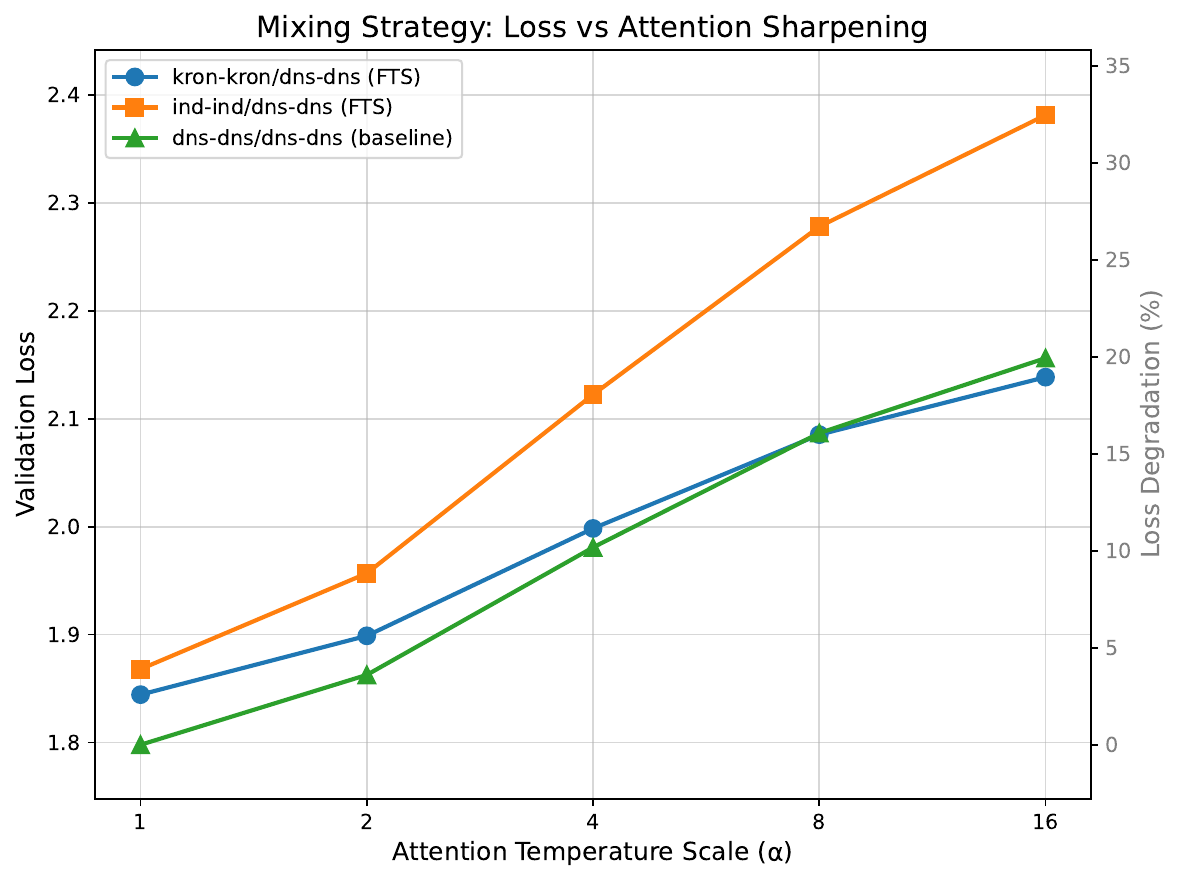} &
\includegraphics[width=0.48\textwidth]{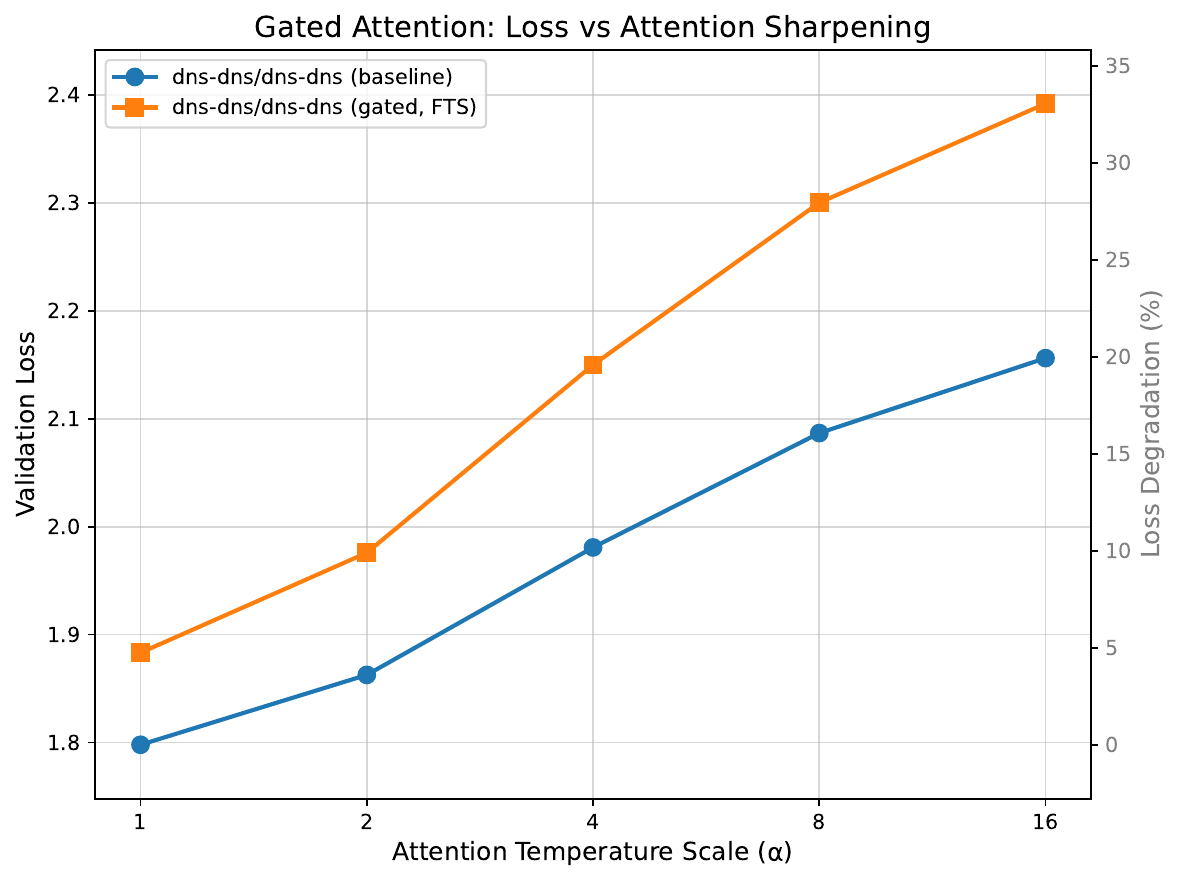} \\
(a) Mixing strategy comparison & (b) Gated attention comparison \\
\end{tabular}
\caption{\textbf{Validation loss versus attention amplification factor.} (a) Kronecker-Dense mixing degrades more gracefully than Independent-Dense, with 9.5\% lower cumulative degradation across $\alpha \in [1,16]$. (b) Gated attention variants show similar amplification robustness to their non-gated counterparts. All configurations maintain functional generation at $\alpha=16$ with bounded degradation.}
\label{fig:alpha_degradation}
\end{figure}

Figure~\ref{fig:alpha_degradation} visualizes these degradation curves, showing that Kronecker mixing maintains near-linear growth in loss as $\alpha$ increases, while Independent mixing exhibits steeper acceleration at higher amplification factors.

\subsection{Kronecker Communication Structure}
\label{sec:kronecker_structure}

The $H \times H$ Kronecker mixing matrices provide explicit routing weights between attention heads. Unlike dense mixing where cross-head communication is implicit within high-dimensional projections, Kronecker mixing exposes head-to-head information flow as scalar weights that can be directly inspected. We visualize these learned routing patterns to understand how heads coordinate information exchange.

\begin{figure}[!htb]
\centering
\begin{tabular}{c}
\includegraphics[width=0.95\textwidth]{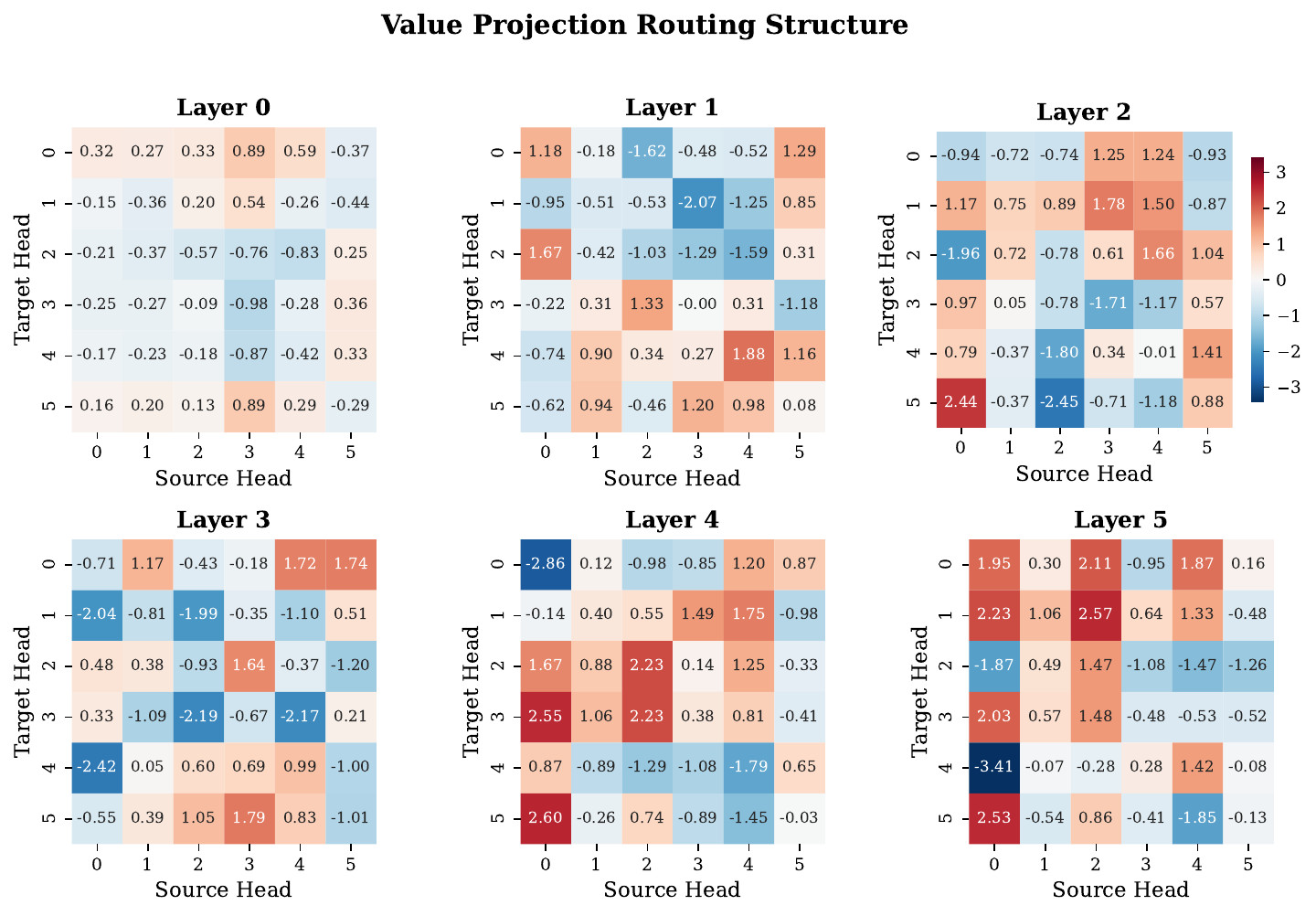} \\
(a) Value Projection Routing \\[2mm]
\includegraphics[width=0.95\textwidth]{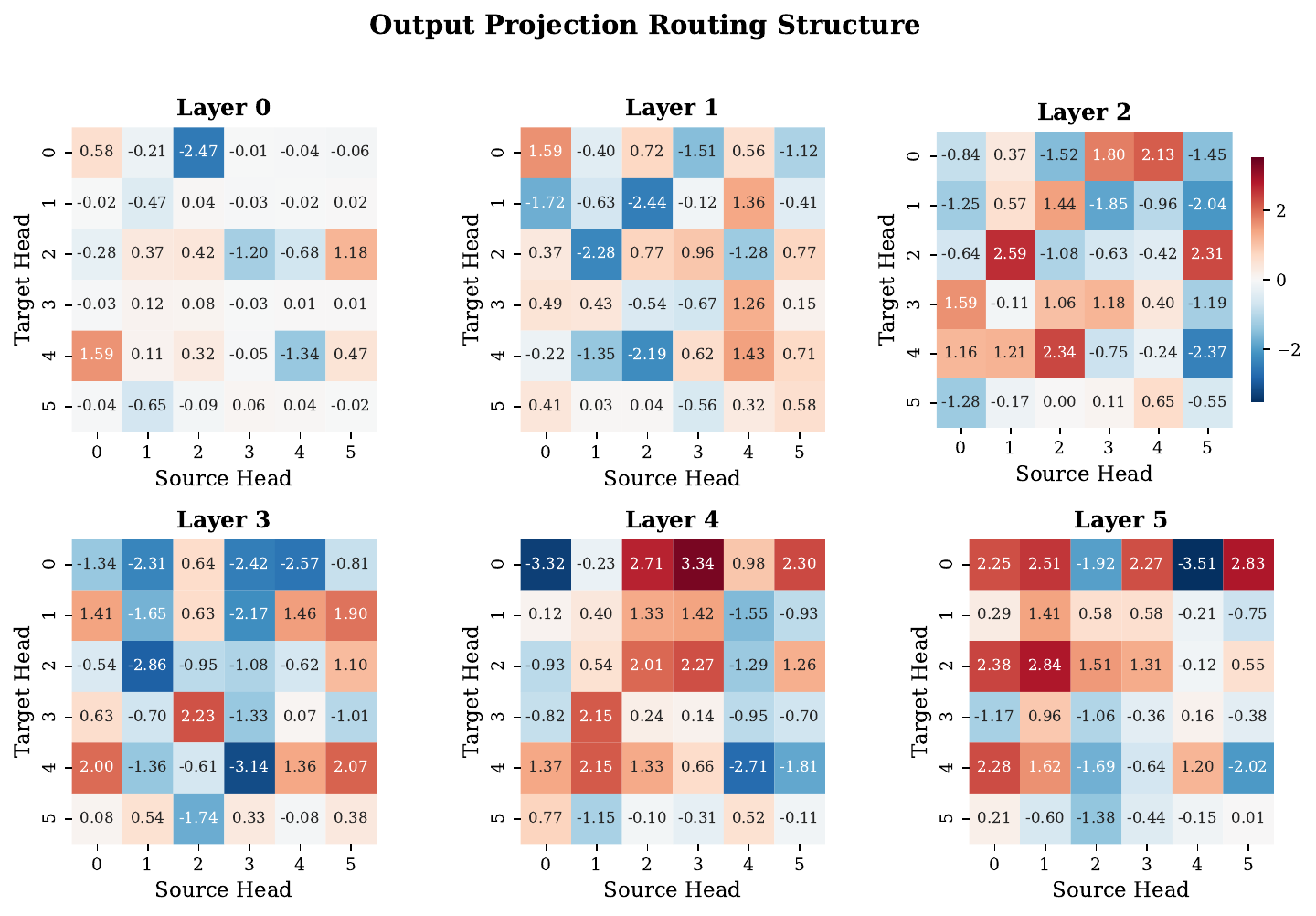} \\
(b) Output Projection Routing \\
\end{tabular}
\caption{\textbf{Learned Kronecker routing matrices across layers.} Each heatmap shows the $6 \times 6$ head-to-head routing weights for (a) value projection and (b) output projection. Cell $(i,j)$ indicates routing from head $j$ to head $i$. Red indicates positive (excitatory) weights; blue indicates negative (inhibitory) weights. Routing strength increases in deeper layers, with maximum weights growing from $\sim$1.0 in layer 0 to $\sim$3.5 in layer 5.}
\label{fig:kronecker_routing}
\end{figure}

Figure~\ref{fig:kronecker_routing} shows the routing matrices for value (V) and output (O) projections across all six layers. Each cell $(i,j)$ represents the scalar weight controlling information flow from head $j$ to head $i$. Positive weights (red) indicate excitatory routing where source head content is amplified; negative weights (blue) indicate inhibitory routing where source content is suppressed or inverted.

\paragraph{Layer-wise Evolution.} The routing matrices reveal systematic evolution across the network depth. Early layers (0-1) show relatively uniform weights near unity, suggesting broad information sharing during initial feature extraction. Middle layers (2-3) develop more structured patterns with specific head pairs forming strong connections while others attenuate. Deep layers (4-5) exhibit the strongest routing weights (up to 3.5$\times$ amplification), indicating that heads in later layers rely heavily on coordinated information from specific partners.

\paragraph{Hub Structure.} Head 0 emerges as a routing hub in both V and O projections, serving as a central nexus for information aggregation. This hub receives strong inputs from multiple source heads and broadcasts widely to targets. The hub structure suggests hierarchical organization where certain heads specialize in integrating information from specialists, potentially enabling compositional computation where syntactic and semantic features combine through the hub.

\paragraph{Connection to Amplification Robustness.} The routing structure explains why Kronecker mixing degrades more gracefully under attention amplification (Table~\ref{tab:amplification}). When attention distributions sharpen, individual heads may make suboptimal discrete selections. The learned routing weights allow complementary heads to compensate: if head 3 makes a poor selection, heads 0 and 4 (which receive its output with specific learned weights) can route around the error through their own selections. Independent mixing lacks this compensation mechanism, explaining its 27\% degradation versus Kronecker's 16\%.

\subsection{Attention Pattern Visualization}

\begin{figure}[h]
\centering
\begin{tabular}{cccc}
\includegraphics[width=0.22\textwidth]{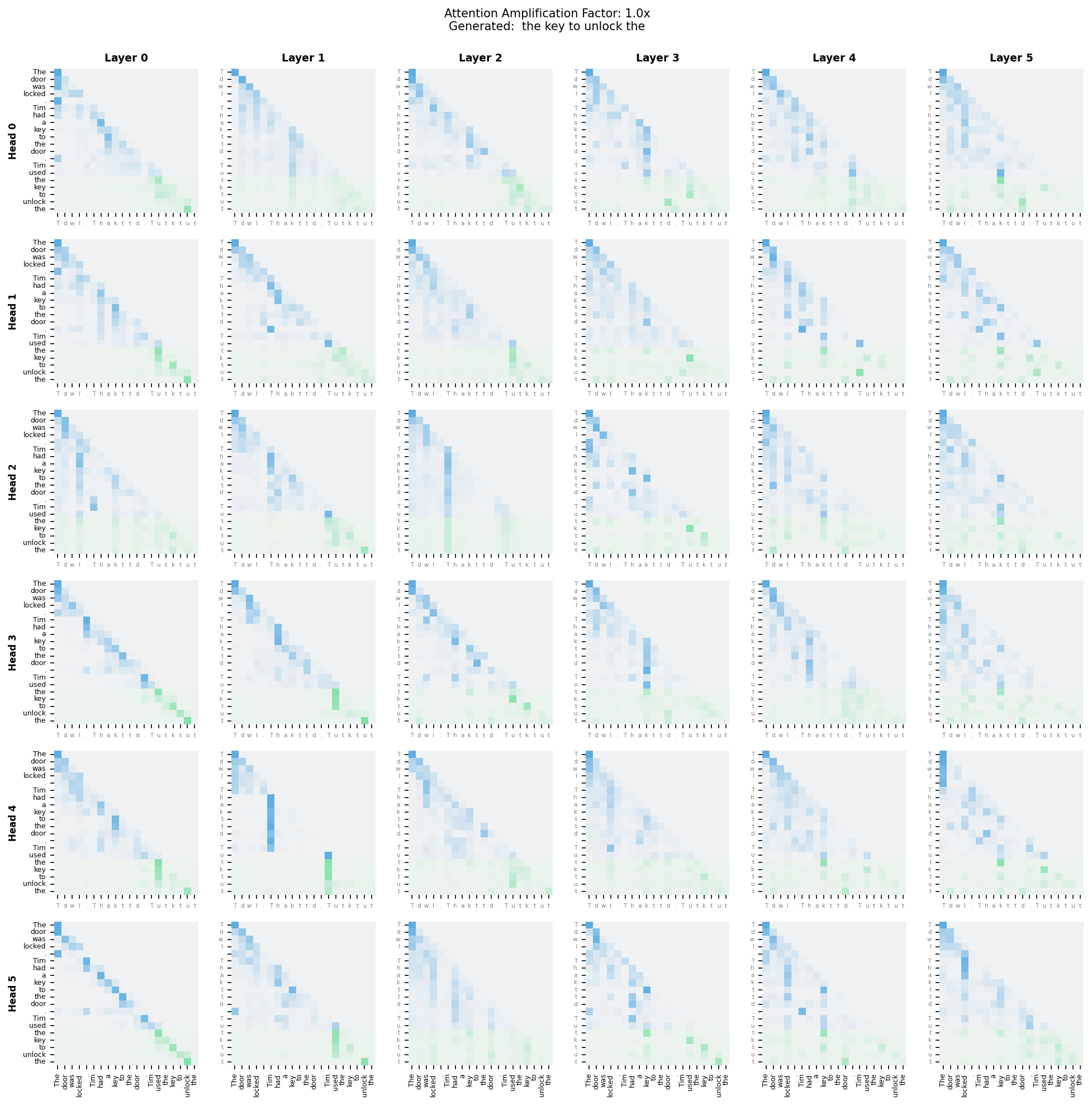} &
\includegraphics[width=0.22\textwidth]{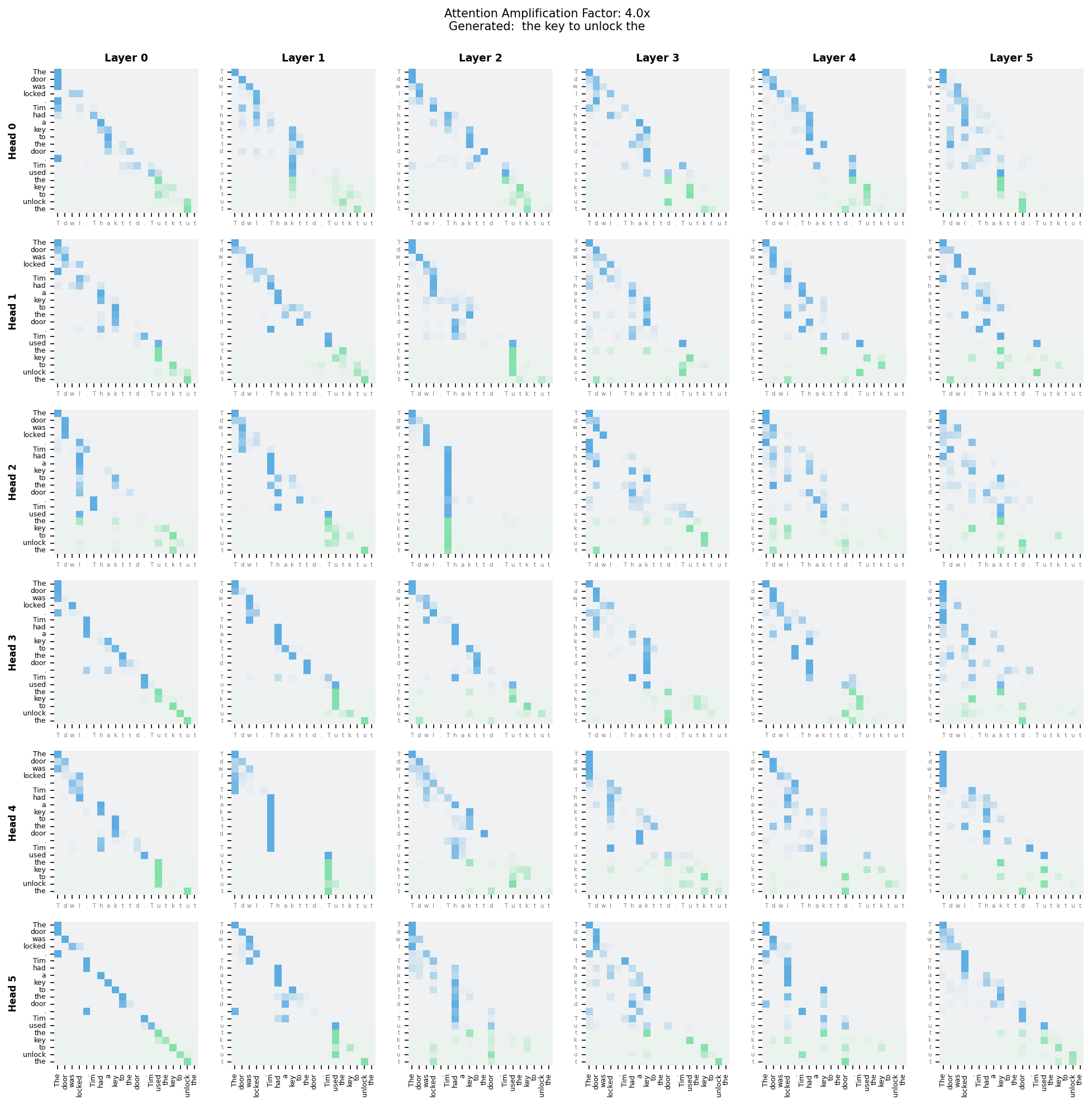} &
\includegraphics[width=0.22\textwidth]{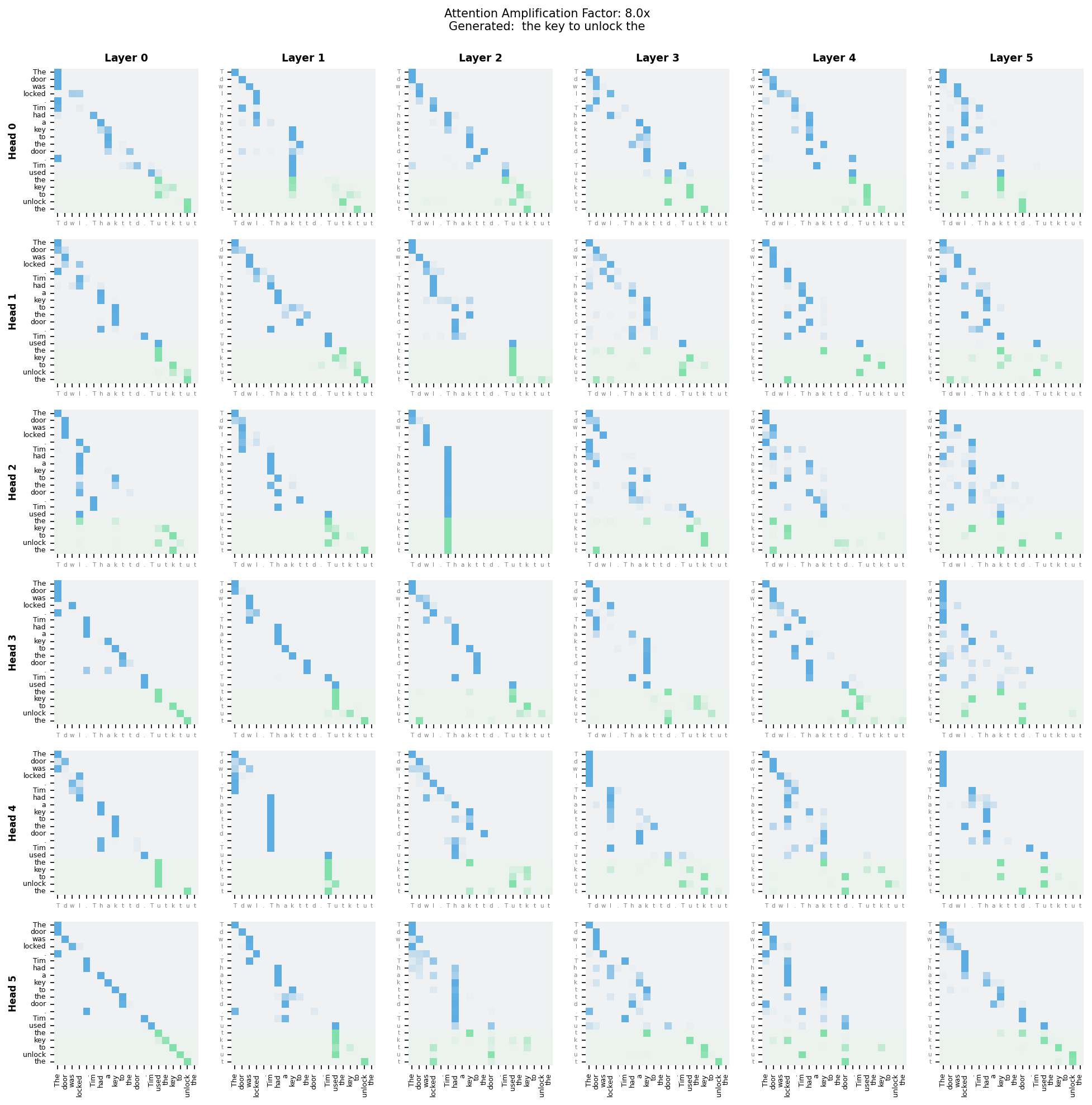} &
\includegraphics[width=0.22\textwidth]{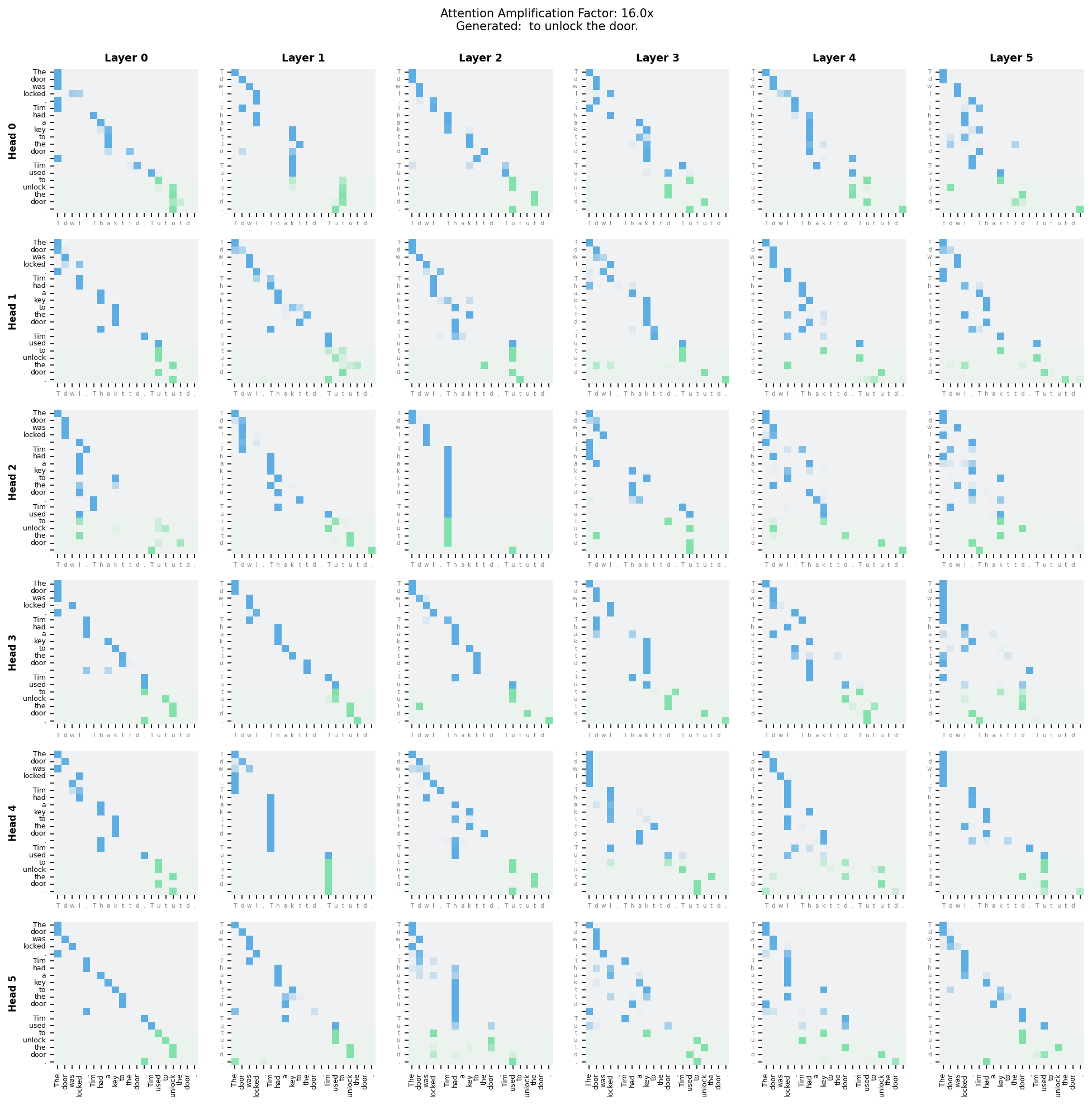} \\
$\alpha=1$ & $\alpha=4$ & $\alpha=8$ & $\alpha=16$ \\
\end{tabular}
\caption{\textbf{Attention patterns under progressive amplification (FTS configuration, Layer 3).} Distributions sharpen from soft mixing at baseline to near-deterministic selection at $\alpha=16$. The model maintains coherent predictions throughout this transition, indicating that the underlying computation can operate on discrete token selections.}
\label{fig:attention}
\end{figure}

Figure~\ref{fig:attention} shows attention heatmaps from a single evaluation sample across amplification factors, revealing the transition from distributed mixing to focused selection. At $\alpha=1$ (baseline training condition), attention spreads across multiple tokens with relatively uniform weights. At $\alpha=4$, patterns begin to concentrate on specific positions while retaining some distribution. At $\alpha=8$, attention becomes highly focused with clear semantic structure. At $\alpha=16$, distributions approach binary selection where one token dominates each query position.

The preservation of semantic structure during sharpening is notable. At $\alpha=8$, attention concentrates on syntactically and semantically relevant positions rather than collapsing to arbitrary nearest neighbors or position-based patterns. This suggests the attention patterns reflect learned dependencies in the data rather than artifacts of the softmax operation. At $\alpha=16$, while distributions become near-binary, the selected tokens correspond to meaningful relationships in the input sequence.

\subsection{Head Specialization Patterns}

We compare head specialization between architectures by measuring each head's performance on coreference resolution, a task that requires identifying which entity a pronoun or definite description refers to. For each layer-head combination, we compute the percentage of evaluation samples where the head attends most strongly to the correct antecedent token.

\begin{figure}[h]
\centering
\includegraphics[width=0.95\textwidth]{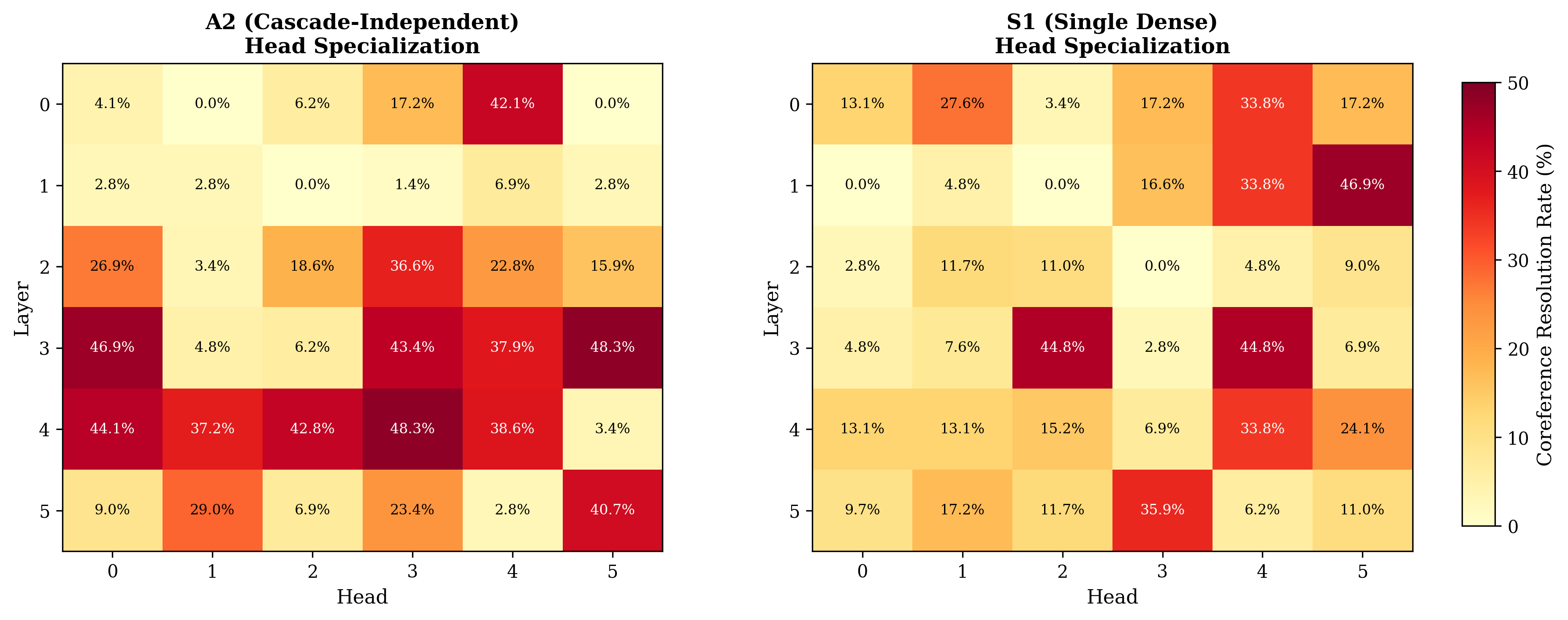}
\caption{\textbf{Coreference resolution accuracy by head across architectures.} Left: Independent-Dense configuration shows strong specialization with distinct functional roles per head. Right: Dense baseline distributes coreference computation across multiple heads with less differentiation. Architectural constraints promote interpretable specialization.}
\label{fig:specialization}
\end{figure}

Figure~\ref{fig:specialization} compares specialization patterns between Independent-Dense (maximum architectural constraint) and Dense baseline configurations. Independent-Dense shows stronger differentiation between heads, with clear specialists emerging. For example, Layer 4 Head 3 achieves 48.3\% coreference accuracy compared to a distributed 25-30\% across heads in the Dense baseline. This indicates that architectural constraints encourage functional specialization, where different heads learn to perform distinct operations rather than redundantly computing similar functions. This observation aligns with recent characterizations of attention heads serving diverse cognitive roles \cite{ma2025cognitive} and the measurement of head differentiation through local learning dynamics \cite{wang2024rllc}.

The emergence of specialists in channelized architectures has implications for interpretability \cite{olsson2022context}. When heads perform distinct functions, interventions can be targeted precisely \cite{meng2022locating}. When computation is distributed across redundant heads, isolating the causal role of any single component becomes difficult. The tradeoff is that distributed computation provides robustness through redundancy, while specialized computation provides interpretability through isolation. The Kronecker routing structure (Section~\ref{sec:kronecker_structure}) reveals how these specialists coordinate: coreference-specialized heads in later layers receive strong routing weights from syntactic heads in earlier layers, suggesting hierarchical information flow from structural analysis to reference resolution.

\section{Discussion}

\subsection{Robustness Under Discretization}

The attention amplification experiments reveal that all configurations maintain functional generation when attention distributions are sharpened to near-deterministic selection. This robustness is notable because standard transformer intuitions suggest that models learn distributed representations where meaning emerges from weighted combinations of features. Forcing hard selection from such representations should cause catastrophic failure as the model loses access to the nuanced mixing ratios it relied on during training.

The observed degradation of 16-27\% represents a change in calibration rather than computational collapse. Models continue to generate semantically appropriate continuations and maintain syntactic coherence at $\alpha=16$, suggesting the underlying selection mechanisms remain intact. The loss increase appears to stem from over-confidence in predictions rather than failure to identify relevant tokens. This pattern is consistent with models that learn pointer-based selection algorithms during training but apply soft probabilistic smoothing at inference time for uncertainty quantification. The preservation of functionality under discretization connects to recent work on latent reasoning in language models \cite{hao2024coconut,liu2025jepa,ning2025latent}, which demonstrates that transformers can perform multi-step reasoning through internal computation rather than explicit soft mixing.

The superior amplification robustness of Kronecker mixing (16\% degradation) compared to Independent mixing (27\% degradation) provides insight into how architectural constraints shape learned computation. Kronecker mixing permits scalar communication between heads through the $H \times H$ mixing matrix, allowing heads to coordinate their selections as distributions sharpen. The routing structure analysis (Section~\ref{sec:kronecker_structure}, Figure~\ref{fig:kronecker_routing}) reveals that these learned weights form hub-and-spoke patterns where central heads aggregate information from specialists, enabling error compensation when individual heads make poor discrete selections. Independent mixing isolates heads completely, preventing this compensation mechanism. The 9.5\% cumulative advantage for Kronecker suggests that interpretable architectures need not completely eliminate cross-head communication, but rather should constrain it to preserve inspectability. This aligns with recent work demonstrating parameter efficiency advantages of Kronecker factorizations \cite{sadeghi2025moka}.

\subsection{Design Recommendations}

The experimental results establish three configurations that balance interpretability and performance requirements differently. For applications requiring maximum transparency into model computation, Frozen-Token-Stream mode with Fully Independent mixing (\texttt{ind-ind/ind-ind}) provides complete isolation of head functions and pure token embeddings in the token stream. This configuration incurs an 8\% validation loss cost but enables surgical analysis where each head's contribution can be measured independently. Attention amplification at $\alpha=8$ further reveals discrete algorithmic structure for inspection.

For applications requiring interpretability with minimal performance sacrifice, Frozen-Token-Stream mode with Kronecker-Dense mixing (\texttt{kron-kron/dns-dns}) provides the recommended configuration. The 2.5\% validation loss cost is offset by the $H \times H$ mixing matrices that make cross-head communication explicit and inspectable, as demonstrated in Figure~\ref{fig:kronecker_routing}. The scalar routing weights provide interpretable coordination between heads while maintaining near-baseline performance. This configuration serves most interpretability applications where modest performance costs are acceptable.

For applications prioritizing performance but desiring architectural support for analysis, Token-Factor mode with Dense mixing (\texttt{dns-dns/dns-dns}) provides standard transformer behavior while maintaining the dual-stream infrastructure. The stream decomposition enables ablation studies and attention amplification analysis even without channelized mixing constraints. This configuration allows practitioners to develop models with conventional optimization characteristics while retaining architectural hooks for interpretability analysis.

Head count should be selected based on the desired granularity of specialization. The head scaling experiments demonstrate that increasing heads from 4 to 16 improves specialization from 0.42 to 0.85 while modestly improving performance. For interpretability applications, we recommend at least 8 heads to enable meaningful functional differentiation. Head dimension should not fall below 32 to avoid capacity bottlenecks, though our experiments with $d_h=64$ for 8 heads showed no degradation relative to fewer heads with larger dimensions.

\subsection{Limitations and Future Work}

Our experiments use models at 29M parameters trained on a pedagogical corpus of grade-school instructional materials. The controlled nature of this data likely makes interpretable structure easier to learn than would occur with web-scale pretraining on diverse internet text. Whether the interpretability-performance tradeoffs scale favorably to billion-parameter models trained on noisy data requires investigation. The bounded costs we observe (2.5-8\%) may increase at larger scales if distributed computation becomes more critical for handling diverse linguistic phenomena.

The attention amplification analysis was conducted at inference time on models trained with standard $\alpha=1$ softmax. Whether training with progressive amplification schedules could further encourage discrete algorithmic structure remains unexplored. Such training might reduce the baseline-to-amplified performance gap, making models more robust to inspection under sharpening.

Our analysis focuses on language modeling loss and limited probing of head specialization. Comprehensive evaluation should include downstream tasks, human evaluation of generated text quality under various configurations, and detailed circuit analysis identifying which heads implement which specific operations. The head specialization analysis (Figure~\ref{fig:specialization}) demonstrates functional differentiation exists, but complete taxonomies of learned operations across heads would strengthen interpretability claims.

The architecture provides infrastructure for interpretability but does not automatically make all model behavior transparent. Attention patterns under amplification reveal which tokens influence predictions, but understanding why those selections occur and what computations the FFN performs on selected tokens requires additional analysis \cite{bricken2023monosemanticity}. The dual-stream decomposition and channelized mixing create necessary conditions for interpretability by isolating computation into inspectable units, but sufficient understanding requires complementary analysis methods \cite{belinkov2022probing}.

\subsection{Comparison to Alternative Approaches}

Several recent architectures pursue interpretability through different mechanisms. Mixture of Experts models \cite{shazeer2017outrageously,fedus2021switch} achieve modularity through explicit routing but require non-differentiable selection mechanisms and typically route entire tokens rather than feature dimensions. Our channelized mixing provides differentiable modular computation within each layer while maintaining gradient flow through all heads. DeBERTa \cite{he2021deberta} separates position and content computation but allows immediate mixing in downstream layers, while our dual-stream architecture maintains persistent separation through all layers until final output integration.

The circuits literature \cite{elhage2021mathematical,olsson2022context} demonstrates that interpretable computational patterns emerge in standard transformers, showing that architectural constraints are not strictly necessary for interpretability. However, the difficulty of circuit discovery at scale and the tendency of models to develop redundant computational pathways suggests that architectures designed for interpretability may scale more effectively than post-hoc analysis of unconstrained models. Related work on modular training \cite{liu2024seeing} shows that locality constraints can encourage interpretable structure during learning. Our results show that modest architectural constraints can encourage specialized computational patterns while incurring bounded performance costs, potentially making interpretability more tractable at larger scales.

\section{Related Work}

\subsection{Mechanistic Interpretability}

The circuits framework \cite{elhage2021mathematical,olah2020zoom} established that transformer computations can be understood as compositions of interpretable circuits operating on a shared residual stream. This view frames the residual stream as a communication channel where attention heads read from and write to specific subspaces. Olsson et al. \cite{olsson2022context} identified induction heads as a canonical example of such circuits, demonstrating that specific attention patterns implement algorithmic copying behavior. Wang et al. \cite{wang2023interpretability} extended this analysis to indirect object identification in GPT-2, revealing multi-step reasoning circuits spanning multiple layers.

Our architecture builds on this foundation by constraining the residual stream structure to make circuit identification more tractable. Where standard transformers require careful analysis to disentangle overlapping computations \cite{elhage2022superposition}, our channelized mixing enforces separation by design. The dual-stream decomposition further isolates token-derived computations from contextual transformations, reducing the search space for circuit discovery.

\subsection{Attention Analysis and Interpretation}

Whether attention weights provide faithful explanations of model behavior remains contested. Jain and Wallace \cite{jain2019attention} demonstrated that attention distributions can be dramatically altered without changing predictions, questioning their explanatory value. Wiegreffe and Pinter \cite{wiegreffe2019attention} countered that this depends on the definition of explanation, showing that attention patterns correlate with input importance under appropriate conditions.

Empirical analyses reveal structured attention behavior in pretrained models. Clark et al. \cite{clark2019does} found that BERT attention heads specialize for syntactic relations including coreference, dependency parsing, and named entity recognition. Voita et al. \cite{voita2019analyzing} showed that specialized heads perform most of the useful computation while others can be pruned with minimal impact. Kovaleva et al. \cite{kovaleva2019revealing} identified common attention patterns including diagonal, vertical, and block structures across BERT layers.

Our architecture addresses the attention-as-explanation debate by making attention's causal role more transparent. Stream separation ensures that attention output flows exclusively to the token stream, isolating its contribution from FFN effects. Channelized mixing prevents attention computations from being distributed across redundant pathways. The attention amplification analysis (Section~\ref{sec:amplification}) further probes whether models rely on soft weighted combinations or discrete token selection.

\subsection{Disentangled Representations}

DeBERTa \cite{he2021deberta} introduced disentangled attention where content and position information are computed separately then combined additively. This improves performance on tasks requiring fine-grained positional reasoning but allows immediate mixing in downstream layers. Position encoding methods including ALiBi \cite{press2022train} and RoPE \cite{su2021roformer} provide alternative approaches to incorporating positional information \cite{dufter2022position}.

Our dual-stream architecture pursues a different form of disentanglement: separating token-derived information from contextual transformations rather than content from position. The token stream preserves information derived directly from input token identities, while the context stream accumulates learned transformations. This separation is maintained through all layers until final output integration, enabling surgical analysis of each stream's contribution through ablation experiments.

\subsection{Modular and Sparse Architectures}

Mixture of Experts (MoE) architectures \cite{shazeer2017outrageously,fedus2021switch} achieve modularity through explicit routing, where each token is processed by a subset of expert networks. This provides computational efficiency and implicit specialization but requires non-differentiable selection mechanisms. Neural module networks \cite{andreas2016neural} compose specialized modules based on input structure, enabling interpretable reasoning for visual question answering.

Sparse attention patterns \cite{child2019generating} reduce computational cost while potentially improving interpretability by limiting which positions can attend to each other. However, sparsity patterns are typically fixed or learned globally rather than adapted per-input.

Our channelized mixing provides a different form of modularity: continuous, differentiable constraints on information flow between attention heads. The Kronecker mixing strategy permits scalar communication between heads while preserving within-head structure, achieving interpretable coordination without explicit routing decisions. This implicit modularity maintains gradient flow through all components while encouraging functional specialization.

\subsection{Probing and Attribution Methods}

Probing classifiers \cite{tenney2019bert,belinkov2022probing} assess what information is encoded in neural representations by training auxiliary classifiers on intermediate activations. This reveals that transformer layers progressively encode syntactic and semantic information, with lower layers capturing surface features and higher layers capturing abstract relations. However, probing measures correlation rather than causation, and high probe accuracy may reflect information that models do not actually use for prediction.

Attribution methods including integrated gradients \cite{sundararajan2017axiomatic} assign importance scores to input features based on their contribution to outputs. These provide instance-level explanations but require careful interpretation and may not reveal the computational mechanisms underlying predictions.

Our approach complements these methods by providing architectural support for causal analysis. Stream ablation experiments directly measure each component's contribution by removing or corrupting it at inference time. The persistent separation of streams enables this surgical analysis without the confounds introduced by entangled representations.

\subsection{Sparse Autoencoders and Feature Decomposition}

Recent work uses sparse autoencoders to decompose neural network activations into interpretable features \cite{cunningham2023sparse,bricken2023monosemanticity}. These methods train auxiliary models to reconstruct activations using sparse combinations of learned feature directions, revealing monosemantic units that correspond to interpretable concepts. Elhage et al. \cite{elhage2022superposition} showed that neural networks represent more features than dimensions through superposition, explaining why individual neurons often exhibit polysemantic behavior.

Geva et al. \cite{geva2021transformer} demonstrated that feed-forward layers function as key-value memories, with individual neurons responding to specific input patterns and contributing predictable outputs. This view suggests FFN computation may be more structured than previously assumed.

Our architecture does not directly address superposition or feature decomposition, but the channelized constraints may reduce the need for such post-hoc analysis by encouraging cleaner separation during training. The head specialization patterns we observe (Figure~\ref{fig:specialization}) suggest that architectural constraints can promote functional differentiation without auxiliary decomposition methods.

\subsection{Architectural Design for Interpretability}

Liu et al. \cite{liu2024seeing} proposed brain-inspired modular training (BIMT) using locality constraints to encourage interpretable structure. Their approach penalizes long-range connections in the computation graph, promoting modular organization that mirrors biological neural networks.

Our work shares the goal of designing for interpretability rather than excavating it post-hoc, but pursues different mechanisms. Where BIMT constrains spatial connectivity, we constrain information flow topology through mixing strategies and stream separation. The channelized hierarchy (Identity $\subset$ Independent $\subset$ Kronecker $\subset$ Dense) provides practitioners with explicit control over the interpretability-performance tradeoff, enabling application-specific configuration rather than fixed architectural choices.

\section{Conclusion}

We introduced the Dual-Stream Transformer, an architecture that enforces interpretable structure through dual-stream decomposition and channelized mixing. The residual stream is factored into token and context components with functional separation: attention updates the token stream while feed-forward networks update the context stream. Information flow between attention heads is controlled through a hierarchy of mixing strategies ranging from fully independent to dense, exposing a tunable interpretability-performance tradeoff.

Experiments at 29M parameters on language modeling tasks demonstrate that interpretability costs are bounded and predictable. The recommended Kronecker mixing configuration costs 2.5\% validation loss while making cross-head communication explicit through learned scalar weights. Fully independent mixing provides maximum interpretability at 8\% cost. All configurations maintain functional generation under attention amplification up to 16x sharpening with degradation ranging from 16\% to 27\%, suggesting learned computation operates on discrete token selections even when trained with soft probabilistic distributions.

The architecture provides practitioners with explicit configuration choices based on application requirements. Safety-critical systems requiring transparency can select frozen-token-stream mode with independent mixing, accepting the 8\% performance cost for complete isolation of head functions. Production systems with modest interpretability requirements can use Kronecker mixing at 2.5\% cost. Development and analysis workflows can leverage attention amplification as a diagnostic on any configuration to reveal computational structure.

The dual-stream design demonstrates that interpretability can be an architectural property rather than an emergent phenomenon requiring post-hoc excavation. By constraining information flow topology and maintaining persistent functional separation, the architecture makes internal structure inspectable by design. Future work should investigate whether these architectural constraints scale favorably to billion-parameter models and whether training with progressive attention amplification can further encourage discrete algorithmic structure. Code and trained models are available at \texttt{[URL to be added]}.

\bibliographystyle{plainnat}
\bibliography{references-updated}

\section{Complete Architecture Specification}
\label{app:architecture}

\subsection{Model Hyperparameters}

Table~\ref{tab:hyperparams} lists the hyperparameters for our base model configuration.

\begin{table}[h]
\centering
\caption{Base model hyperparameters.}
\label{tab:hyperparams}
\begin{tabular}{lcc}
\toprule
Hyperparameter & Symbol & Value \\
\midrule
Model dimension & $D$ & 512 \\
Number of layers & $L$ & 6 \\
Number of heads & $H$ & 8 \\
Head dimension & $d_h$ & 64 \\
FFN hidden dimension & $d_{\text{ff}}$ & 2048 \\
Vocabulary size & $V$ & 32,000 \\
Maximum sequence length & $T_{\max}$ & 512 \\
Total parameters & --- & $\sim$29M \\
\bottomrule
\end{tabular}
\end{table}

\subsection{Forward Pass Pseudocode}

Algorithm~\ref{alg:forward} presents the complete forward pass for \textsc{Token-Factor} mode with configurable mixing.

\begin{algorithm}[h]
\caption{Dual-Stream Transformer Forward Pass (\textsc{Token-Factor} mode)}
\label{alg:forward}
\begin{algorithmic}[1]
\REQUIRE Input tokens $\mathbf{t} \in \mathbb{N}^{B \times T}$
\ENSURE Output logits $\mathbf{y} \in \mathbb{R}^{B \times T \times V}$
\STATE $\xt^{(0)} \leftarrow \text{Embedding}(\mathbf{t})$ \COMMENT{Initialize token stream}
\STATE $\xe^{(0)} \leftarrow \mathbf{0}$ \COMMENT{Initialize contextual stream}
\FOR{$\ell = 0$ to $L-1$}
    \STATE $\mathbf{x}^{(\ell)} \leftarrow \xt^{(\ell)} + \xe^{(\ell)}$ \COMMENT{Combined stream}
    \STATE \COMMENT{--- Attention Block ---}
    \STATE $\mathbf{x}_{\text{norm}} \leftarrow \text{CLN}(\mathbf{x}^{(\ell)})$
    \STATE $\mathbf{x}_{t,\text{norm}} \leftarrow \text{CLN}(\xt^{(\ell)})$
    \STATE $Q, K \leftarrow W_Q \mathbf{x}_{\text{norm}}, W_K \mathbf{x}_{\text{norm}}$ \COMMENT{Dense Q/K}
    \STATE $V \leftarrow \text{MixingLinear}_V(\mathbf{x}_{t,\text{norm}})$ \COMMENT{Configurable V mixing}
    \STATE $A \leftarrow \text{softmax}(QK^\top / \sqrt{d_h})$ \COMMENT{Attention weights}
    \STATE $\mathbf{a}_{\text{out}} \leftarrow \text{MixingLinear}_O(A \cdot V)$ \COMMENT{Configurable output mixing}
    \STATE $\xt^{(\ell+1)} \leftarrow \xt^{(\ell)} + \mathbf{a}_{\text{out}}$ \COMMENT{Update token stream}
    \STATE \COMMENT{--- FFN Block ---}
    \STATE $\mathbf{x}^{(\ell+1)} \leftarrow \xt^{(\ell+1)} + \xe^{(\ell)}$ \COMMENT{Recompute combined}
    \STATE $\mathbf{x}_{\text{norm}} \leftarrow \text{CLN}(\mathbf{x}^{(\ell+1)})$
    \STATE $\mathbf{h} \leftarrow \text{GELU}(\text{MixingLinear}_{\text{up}}(\mathbf{x}_{\text{norm}}))$
    \STATE $\mathbf{f}_{\text{out}} \leftarrow \text{MixingLinear}_{\text{down}}(\mathbf{h})$
    \STATE $\xe^{(\ell+1)} \leftarrow \xe^{(\ell)} + \mathbf{f}_{\text{out}}$ \COMMENT{Update contextual stream}
\ENDFOR
\STATE $\mathbf{x}_{\text{final}} \leftarrow \text{LayerNorm}(\xt^{(L)} + \xe^{(L)})$
\STATE $\mathbf{y} \leftarrow W_{\text{lm\_head}} \mathbf{x}_{\text{final}}$
\RETURN $\mathbf{y}$
\end{algorithmic}
\end{algorithm}

\subsection{Mixing Strategy Parameter Counts}

Table~\ref{tab:mixing_params} compares parameter counts for each mixing strategy applied to a $D \times D$ projection with $H$ heads.

\begin{table}[h]
\centering
\caption{Parameter counts for mixing strategies ($D=512$, $H=8$).}
\label{tab:mixing_params}
\begin{tabular}{lcc}
\toprule
Strategy & Parameters & Example ($D=512$, $H=8$) \\
\midrule
\textsc{Identity} & 0 & 0 \\
\textsc{Independent} & $H \cdot d_h^2$ & $8 \times 64^2 = 32,768$ \\
\textsc{Kronecker} & $H^2$ & $8^2 = 64$ \\
\textsc{Dense} & $D^2$ & $512^2 = 262,144$ \\
\bottomrule
\end{tabular}
\end{table}

Note that \textsc{Kronecker} mixing is extremely parameter-efficient ($H^2 = 64$ parameters vs. $D^2 = 262,144$ for Dense), while providing interpretable cross-head communication.

\section{Training Details}
\label{app:training}

\subsection{Optimization}

All models are trained with AdamW optimizer \cite{loshchilov2017decoupled} with the following settings:
\begin{itemize}
    \item Learning rate: $3 \times 10^{-4}$ with cosine annealing to $3 \times 10^{-5}$
    \item Weight decay: 0.1
    \item Betas: $(0.9, 0.95)$
    \item Gradient clipping: 1.0
    \item Warmup steps: 1000
\end{itemize}

\subsection{Training Configuration}

\begin{itemize}
    \item Batch size: 32 sequences
    \item Sequence length: 512 tokens
    \item Training samples: 2M
    \item Epochs: 2
    \item Floating point: Standard precision
    \item Hardware: Single NVIDIA RTX 4090 (24GB)
\end{itemize}

\subsection{Per-Layer Supervision Settings}

When per-layer supervision is enabled:
\begin{itemize}
    \item Base weight $\lambda = 0.1$
    \item Layer weight decay: linear (layer $\ell$ receives weight $\ell / L$)
    \item Applied only during training, not validation
\end{itemize}

\section{Extended Ablations}
\label{app:ablations}

\subsection{Full Mixing Strategy Grid}

Table~\ref{tab:full_mixing} presents validation loss for all tested mixing configurations.

\begin{table}[h]
\centering
\caption{Full mixing strategy ablation (4K vocab, FTS mode, 3 epochs).}
\label{tab:full_mixing}
\begin{tabular}{llcc}
\toprule
Attention Mixing & FFN Mixing & Val Loss & $\Delta$ \\
\midrule
\texttt{dns-dns} & \texttt{dns-dns} & 2.42 & --- \\
\texttt{kron-kron} & \texttt{dns-dns} & 2.48 & +2.5\% \\
\texttt{ind-ind} & \texttt{dns-dns} & 2.50 & +3.3\% \\
\texttt{ind-ind} & \texttt{ind-ind} & 2.62 & +7.9\% \\
\bottomrule
\end{tabular}
\end{table}

\section{Head Specialization Metric}
\label{app:specialization}

The Head Specialization Score (HSS) measures how distinctly heads attend to different patterns:
\begin{equation}
\text{HSS} = \frac{1}{H(H-1)} \sum_{i \neq j} \left(1 - \frac{\mathbf{a}_i^\top \mathbf{a}_j}{\|\mathbf{a}_i\| \|\mathbf{a}_j\|}\right)
\end{equation}
where $\mathbf{a}_h$ is the flattened attention pattern for head $h$ averaged over evaluation samples. HSS $= 0$ indicates identical patterns across all heads; HSS $= 1$ indicates orthogonal patterns.

\end{document}